\documentclass[letterpaper]{article} 
\usepackage{aaai24}  
\usepackage{times}  
\usepackage{helvet}  
\usepackage{courier}  
\usepackage[hyphens]{url}  
\usepackage{graphicx} 
\usepackage{natbib}  
\usepackage{caption} 
\usepackage{algorithm}
\usepackage{algorithmic}
\usepackage{booktabs} 
\usepackage{amsmath}
\usepackage{newfloat}
\usepackage{listings}
\DeclareCaptionStyle{ruled}{labelfont=normalfont,labelsep=colon,strut=off} 
\floatstyle{ruled}
\newfloat{listing}{tb}{lst}{}
\floatname{listing}{Listing}
\title{Mitigating Urban-Rural Disparities in Contrastive Representation Learning\\ with Satellite Imagery}
\author{
    Miao Zhang\textsuperscript{\rm 1}
    Rumi Chunara\textsuperscript{\rm 2}\\
}
\affiliations{
    \textsuperscript{\rm 1}Tandon School of Engineering, New York University\\

    \textsuperscript{\rm 2} Tandon School of Engineering; School of Global Public Health, New York University

    New York, USA\\
    \{miaozhng, rumi.chunara\}@nyu.edu
}

\usepackage{bibentry}

\begin{document}

\maketitle

\begin{abstract}
Satellite imagery is being leveraged for many societally critical tasks across climate, economics, and public health. Yet, because of heterogeneity in landscapes (e.g. how a road looks in different places), models can show disparate performance across geographic areas. Given the important potential of disparities in algorithmic systems used in societal contexts, here we consider the risk of urban-rural disparities in identification of land-cover features. This is via semantic segmentation (a common computer vision task in which image regions are labelled according to what is being shown) which uses pre-trained image representations generated via contrastive self-supervised learning. We propose fair dense representation with contrastive learning (FairDCL) as a method for de-biasing the multi-level latent space of convolution neural network models. The method improves feature identification by removing spurious model representations which are disparately distributed across urban and rural areas, and is achieved in an unsupervised way by contrastive pre-training. The obtained image representation mitigates downstream urban-rural prediction disparities and outperforms state-of-the-art baselines on real-world satellite images. Embedding space evaluation and ablation studies further demonstrate FairDCL's robustness. As generalizability and robustness in geographic imagery is a nascent topic, our work motivates researchers to consider metrics beyond average accuracy in such applications.
\end{abstract}

\section{Introduction}
\label{sec:intro}
Dense pixel-level image recognition via deep learning for tasks such as segmentation have a variety of applications in landscape feature analysis from satellite images. For example, regional water quality analysis~\cite{griffith2002geographic} or dust emission estimation~\cite{von2019assessing}. Success of the methods rely on powerful visual representations that include both local and global information. However, since pixel-level annotations are usually costly, fully supervised learning is challenging when the amount and variety of labeled data is scarce. Therefore, self-supervised learning is a promising alternative via pre-training a image encoder and transferring learnt representations to downstream problems. As a mainstream, contrastive self-supervised techniques have shown state-of-the-art performance in learning image representations for land cover semantic segmentation across locations~\cite{ayush2021geography, scheibenreif2022self}. In particular, as labeled images are hard to obtain for satellite images, and contrastive approaches do not require labels, they have demonstrated benefits in many real-world tasks including monitoring dynamic land surface~\cite{saha2020unsupervised},  irrigation detection from uncurated and unlabeled satellite images~\cite{agastya2021self}, and volcanic unrest detection with scarce image label~\cite{bountos2021self}.

Importantly, recent attention in machine learning systems has highlighted performance inequities including those by geographic area \cite{Vries_2019_CVPR_Workshops, xie2022fairness, setianto2021spatial,majumdar2022detecting,aiken2023fairness}, and how prediction inequities would compromise policy-making goals~\cite{kondmann2021under}. Disparities at a geographic level fall into the fairness literature due to implications of unequal distribution of resources, opportunities, and essential services, leading to disparities in quality of life and opportunities for those who live in specific areas \cite{hay1995concepts}. As recent work has reinforced, disparities of machine learning model prediction at a geographic level often show disparate performance with respect to minoritized groups or already under-resourced areas \cite{kondmann2021under}. Therefore, given the increased potential of self-supervised contrastive learning, here we turn attention to disparity risks in recognition outcomes between urban and rural areas. The consequences of such recognition tasks have wide usage for societal decisions including urban planning, climate change and disaster risk assessment  ~\cite{mehrabi2021survey, soden2019taking}, so disparity shapes an important concern. Further, while recent work has identified disparities with satellite image representation, specifically across urban and rural lines, and shown the negative consequence on poverty prediction~\cite{aiken2023fairness}, there is limited work on mitigating urban-rural disparities with state-of-the-art vision recognition schemes for landscape analysis, despite their wide applications. 

To bridge this gap, we examine the task of land-cover segmentation and identify disparities across urban and rural areas on satellite images from different locations. As previous work shows, segmentation performance can be disparate across geography types. For example, in areas where land-cover objects have higher density or heterogeneity, performance will be lower even for similar training sample sizes~\cite{zhang2022segmenting}. 
Moreover, identifying and thus addressing disparities for geographic object segmentation is different from classification tasks in other image types such as facial images~\cite{wang2019balanced,ramaswamy2021fair,jung2021fair}. De-biasing classification outcomes relies on robust image-level global representations, which are not ideal for segmentation in which local features are important, thus may not apply to satellite data and relevant tasks. Instead, to our knowledge, we present the first exploration on  learning generalizable and robust local landscape features, while reducing spurious features that are unequally correlated with areas of different urbanization or economic development. (referred to as ``bias'' or ``spurious information''). 
In this way, our work addresses disparity issues in contrastive self-supervised learning for satellite image segmentation.
The specific contributions are: 

\begin{enumerate}
\item We propose a causal model depicting the relationship between landscape features and urban/rural property of images, to unravel the type of implicit bias that a model might learn from data. This framework enables us to identify and address unique disparity challenges in deep learning application for satellite images.

\item For the described bias scenario, we design a fair representation learning method which regularizes the statistical association between pixel-level image features and sensitive variables, termed FairDCL. The methods includes a novel feature map based local mutual information estimation module which incorporates layer-wise fairness regularization into the contrastive optimization objective. Given characteristics of satellite images, this work serves to mitigate performance disparities in downstream landscape segmentation tasks.

\item On real-world satellite datasets, FairDCL shows advantages for learning robust image representation in contrastive pre-training; it surpasses state-of-the-art methods demonstrating smaller urban-rural performance differences and higher worst-case performance, without sacrificing overall accuracy on the target tasks. 

\end{enumerate}

\paragraph{Scope and Limitation.} This work specifically focuses on image representation learning without supervision of labels for the objects to be segmented (also referred to as pre-training), motivated by the approach's effectiveness and low annotation cost as described. Therefore, we do not cover other image analysis schemes, such as supervised or semi-supervised learning and focus on comparison to unsupervised robust representation learning baselines, including gradient reversal learning, domain independent learning, and global representation debiasing with mutual-information. The evaluation of learnt representation quality is achieved by applying a lightweight decoder for the semantic segmentation target to obtain the final downstream task performance, on segmenting common landscape objects, following previous work~\cite{wang2021dense, ziegler2022self}.

\section{Related Work}
\label{sec:related}
\subsection{Self-Supervised Methods for Satellite Images} Semantic segmentation, which quantifies land-cover location and boundary at pixel level, is a fundamental problem in satellite data analysis~\cite{lv2023deep}. Given that per-pixel segmentation annotation required for supervised training is expensive, a growing body of literature leverages self-supervised methods to extract useful image features from large-scale satellite image datasets~\cite{li2021semantic, wang2022self, li2022global}.  Contrastive learning is used as a self-supervised pre-training approach for various downstream vision tasks including classification, detection and segmentation~\cite{chen2020simple, chen2020improved, hendrycks2019using, misra2020self, reed2022self, vu2021medaug, ayush2021geography}. Though most work in this area focuses on optimizing global representations for a single prediction for each image, such as presence of an animal species in an image, ~\cite{wu2018unsupervised, chen2020simple, chen2020improved}, recent work has turned to learning representations suitable for dense predictions (i.e., a prediction for each pixel); such approaches train the model to compare local regions within images, thus preserving pixel-level information~\cite{wang2021dense, o2020unsupervised, chaitanya2020contrastive, xie2021propagate}. Other work~\cite{xiong2020loco} uses overlapped local blocks to increase depth and capacity for decoders that improves local learning. Such methods show the importance of local image representations on dense visual problems like satellite image segmentation, which we leverage for the first time to mitigate disparities on such problems.

When satellite data is collected in multiple temporal resolutions, studies have included contrastive learning methods to learn the representations invariant to subtle landscape variations across the short-term~\cite{mall2023change, ayush2021geography}. However, this type of work requires multi-temporal satellite data and does not consider the same question regarding generalizability with respect to urbanization.


\subsection{Disparities in Image Recognition} Fairness-promoting approaches are being designed in multiple visual recognition domains, generally with human objects and demographic characteristics as the sensitive attributes. For example, in face recognition applications, methods are proposed for mitigating bias across groups like age, gender or race/ethnicity. Such methods include constraining models from learning sensitive information by adversarially training sensitive attribute classifiers ~\cite{raff2018gradient, morales2020sensitivenets}, using penalty losses ~\cite{xu2021consistent, serna2022sensitive}, sensitive information disentanglement~\cite{creager2019flexibly, park2021learning}, and augmenting biased data using generative networks~\cite{ramaswamy2021fair}. Related to healthcare data and practice, methods have shown reduction in bias by altering sensitive features such as skin color but preserve relevant features to the clinical tasks~\cite{yuan2022edgemixup, deng2023fairness}, by augmentation~\cite{burlina2021addressing}, and by adversarial training~\cite{abbasi2020risk, puyol2021fairness}. In comparison, investigation on satellite imagery is limited; Xie \textit{et al.}~\cite{xie2022fairness} formulate disparity among sub-units with linked spatial information, using spatial partitionings instead of sensitive attributes. Aiken \textit{et al.}~\cite{aiken2023fairness} illustrate that urban-rural disparities exist in wealth prediction with satellite images. However, no work has explored disparity in image representation learning for satellite images nor proposed a method to mitigate the same.

A few recent studies have examined robustness and fairness in contrastive learning generally, including an adjusting sampling strategy to restrict models from leveraging sensitive information ~\cite{tsai2021conditional}. However, this approach could lose task-specific information by only letting the model differentiate samples from the same group to avoid learning group boundaries. Two stage training with balanced augmentation~\cite{zhang2022fairness}, fairness-aware losses to penalize sensitive information used in positive and negative pair differentiation~\cite{park2022fair}, and using hard negative samples for contrast to improve representation generalization~\cite{robinson2020contrastive} are other proposed approaches, yet such methods both only apply for classification based on image-level representations opposed to object-level segmentation which is the focus of this work.


\subsection{Summary of Gaps in the Literature}
Existing work in robustness and disparity mitigation for image recognition tasks is limited in multiple ways, with important gaps specifically for satellite imagery. First, some robustness methods assume that spurious feature properties are known, such as skin colors, hair colors, presence of glasses ~\cite{wang2020towards, ramaswamy2021fair, yuan2022edgemixup}, and remove their influences on model performance. The analog of such a property is not available in satellite images, nor is it homogeneous (e.g. each country has unique landscapes relating to urban/rural). Therefore, we do not explicitly define spurious features in our model but automatically extract them with urban/rural discriminators during training. Second, since the existing methods are mostly designed for classification problems, they use image-level representation approaches. However, fairness at an image level would not necessarily extend to pixel-level dense predictions. Third, there is very little work on robust and fair satellite image analysis, for which biased features are harder to discover, interpret and remove, compared to human-object images. Existing work on generalizable satellite representations across temporal changes train models with acquisition date, which is not always available for satellite datasets.


\begin{figure}[h!]
  \centering
    \includegraphics[width=8.4cm]{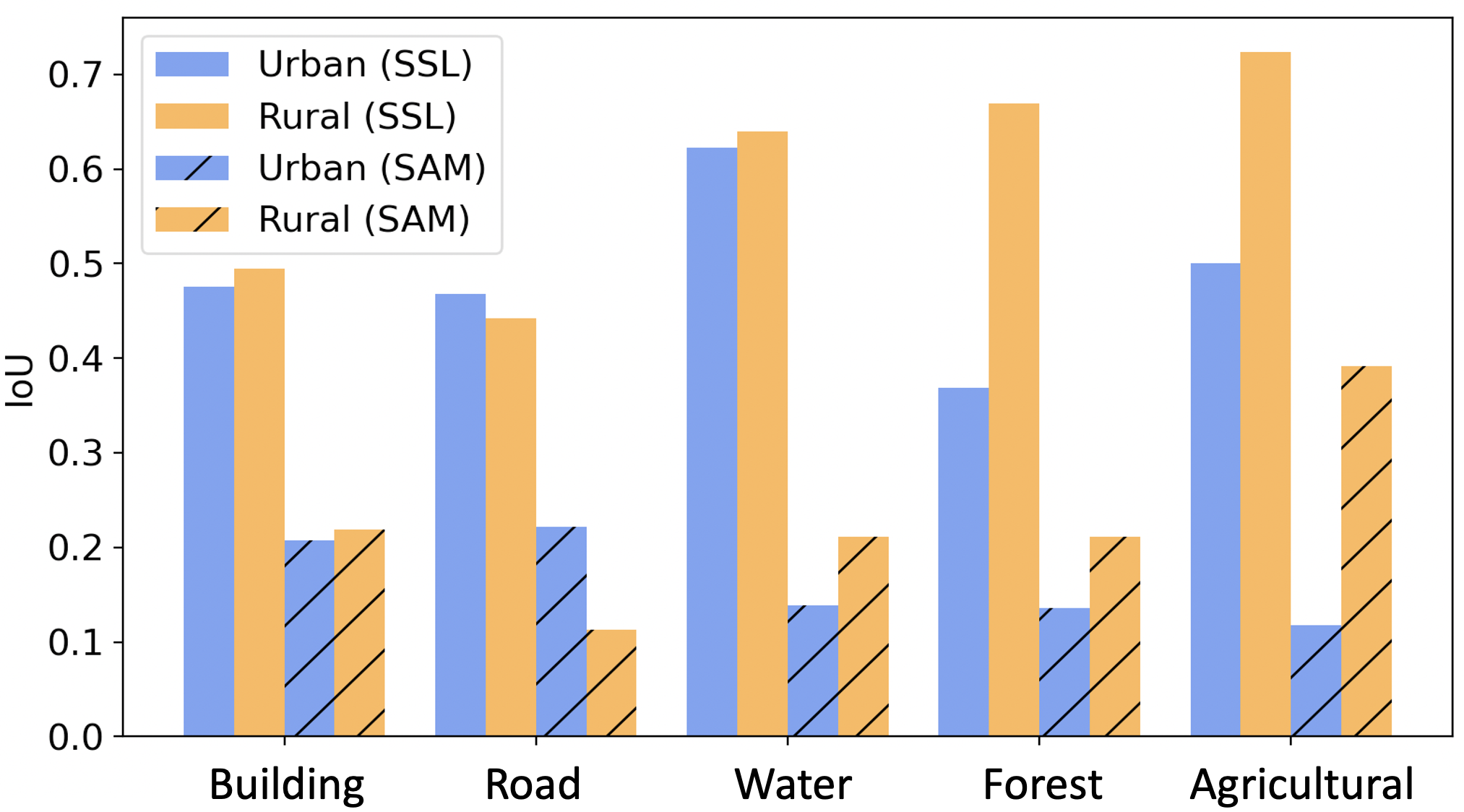}
    \caption{Model segmentation performance on urban and rural images of LoveDA~\cite{wang2021loveda}, measured by intersection-over-union (IoU). Two types of upstream feature encoders are used: (1) CNN encoder trained on unlabeled satellite images with contrastive self-supervised learning (SSL), and (2) pre-trained foundation model Segment Anything (SAM)~\cite{kirillov2023segment}. Urban-rural disparities are observed for land-cover classes with both encoders, and the disadvantaged groups are consistent across learning models.  }
    \label{fig:SAM}
\end{figure}

\section{Problem Statement}
\subsection{Selection of Sensitive Attributes}
In algorithmic fairness studies, sensitive attributes are those that are historically linked to discrimination or bias, and should not be used as the basis for decisions \cite{dwork2012fairness}. Commonly, for example in studies focused on face detection, demographic factors such as race and gender are used as sensitive attributes due to their potential, but unwanted influence on decision-making processes~\cite{lee2018detecting, pessach2022review}. For satellite imagery, while individual-level attributes such as race and gender are not of concern, there are geographical properties such as urban-rural disparities which have precedence both for historical disparities and legal precedence for the need for protection from such disparities \cite{ananian2024employment}. Indeed, rural areas in the United States and globally have lower resources such as health care services \cite{peek2011rural,lin2014microsoft}, higher education disadvantage \cite{roscigno2006education,li2022urban} and lower investment in other areas such as communication technology \cite{nazem1996implementing}. These factors can all significantly impact outcomes for populations in these areas, and it is critical that future decisions impacting rural and urban places do not promulgate such disparities. In terms of operationalizing these attributes, while features of specific urban or rural areas can vary globally, there is consensus that urban areas demarcate cities and their surroundings. Urban areas are very developed, meaning there is a density of human structures, such as houses, commercial buildings, roads, bridges, and railways~\cite{UrbanArea}. In sum, examining urban and rural designations as sensitive attributes can unveil systemic inequalities and aid in creating more equitable algorithms and policies globally.

\subsection{Urban-Rural Disparities with Feature Encoders}
\label{sec:disparity}
We perform satellite image feature extraction with the studied contrastive self-supervised learning (SSL) method, MOCO-v2~\cite{chen2020improved}, and report the semantic segmentation fine-tuning results in Figure~\ref{fig:SAM}. There are several major disparities visible, especially for the class of ``Forest'' and ``Agricultural''. To further expose the issue, we evaluate with a general-purpose feature encoder, Segmenting Anything Model (SAM)~\cite{kirillov2023segment}. It is a vision foundation model trained on a large image dataset (11 millions) of wide geographic coverage for learning comprehensive features. Therefore, the model can transfer zero-shot to image segmentation for our dataset. The results show similar disparities to SSL for each land-cover class (Figure~\ref{fig:SAM}). Motivated by the problem, we propose a causal model to unravel feature relationships in satellite images and the design to utilize robust features to mitigate disparity.

\subsection{Causal Model for Feature Relationships}
\label{sec:problem_unfairness}









\begin{figure}[h!]
  \centering
    \includegraphics[width=8.5cm]{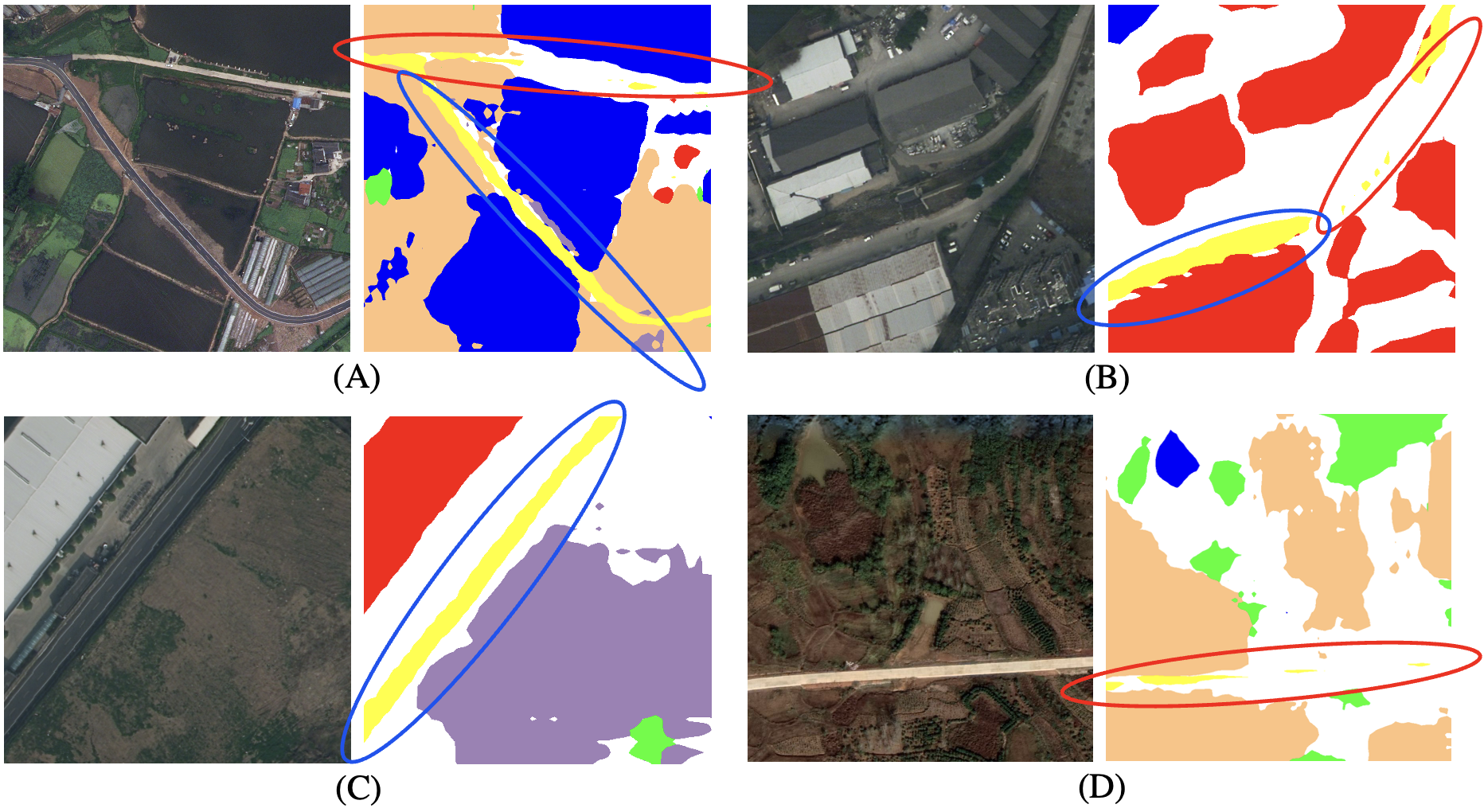}
    \caption{Examples of segmentation bias for ``road'' class due to spurious landscape features; the model segments certain patterns well, like straight and paved road (blue circles), but segments the variations poorly, like curvy and sand road (red circles). }
    \label{fig:unfair}
\end{figure}

Land-cover objects in satellite images, such as residential building, roads, vegetation, etc, often have heterogeneous shapes and distributions in urban and rural areas even within the same geographic region. These distributions are affected by varying levels of development (infrastructure, greening, etc). Considering an attribute $S=\{s_0, s_1\}$ denoting urban/rural area, we define visual representation (high-dimensional embeddings output by model intermediate layers) as $X = \{ X_{spurious}, X_{robust} \}$, where $X_{spurious}$ includes information that varies across urban/rural groups in $S$, for example, the contour, color, or texture of ``road'' or ``building'' class. $X_{robust}$, on the other hand, includes generalizable information, for example, ``road'' segments are narrow and long, while ``building'' segments are clustered. When model output $Y$ is drawn from both $X_{robust}$ and $X_{spurious}$, it can lead to biased performance. For example, roads in grey/blue color (Figure~\ref{fig:unfair} A), with vehicles on them (Figure~\ref{fig:unfair} B), and with lane markings (Figure~\ref{fig:unfair} C), are segmented better (blue circle) than the others (red circle, Figure~\ref{fig:unfair} D). Examples of more classes' spurious and robust components are in the supplementary material A \footnote{Code and supplementary material can be accessed at: https://github.com/ChunaraLab/FairDCL-mitigating-urban-rural-disparity}. 

\label{sec:problem}
\begin{figure}[h!]
  \centering
    \includegraphics[width=5cm]{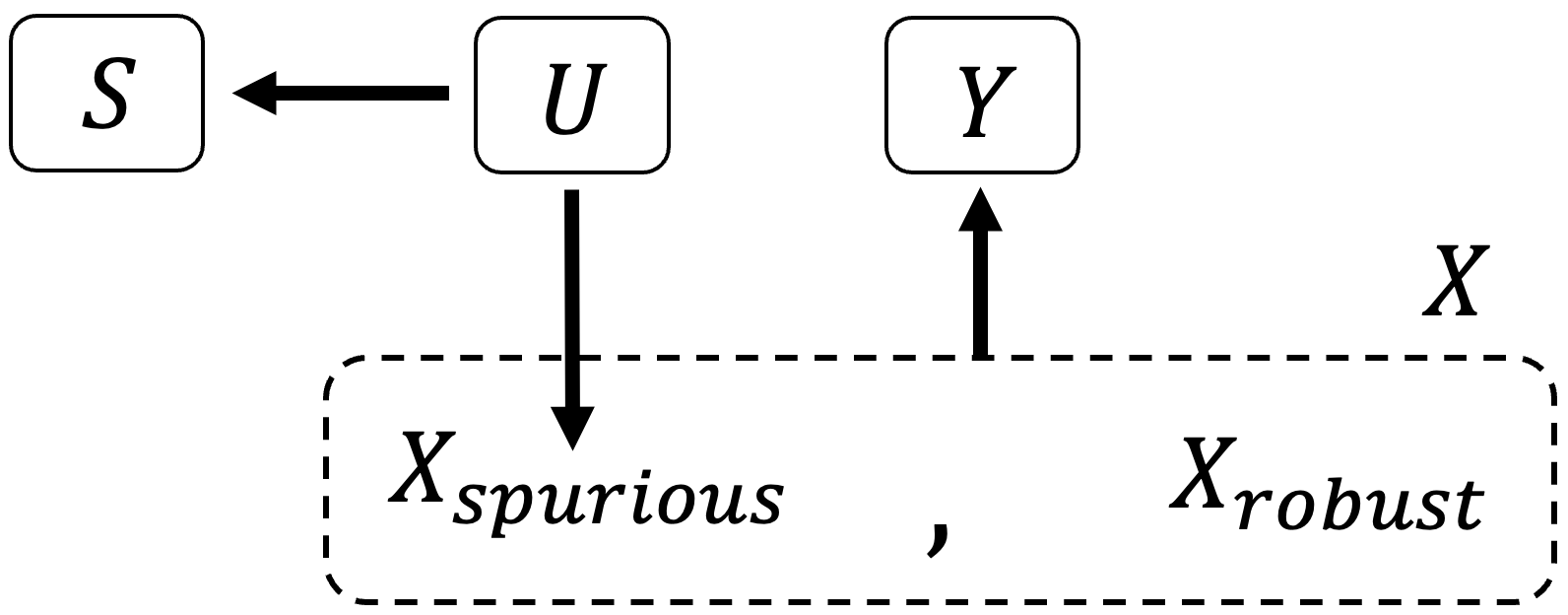}
    \caption{Diagram of defined causal relationships between representation $X$ learnt with contrastive pre-training, target task prediction outputs $Y$, and urban/rural attribute $S$. $X$ contains two parts, $X_{spurious}$ generated from features spuriously correlated to $S$ and $X_{robust}$ generated from independent and unchangeable features. $U$ is unmeasured confounders which cause both $S$ and $X_{spurious}$ thus result in correlations between $S$ and $X_{spurious}$.}
    \label{fig:diagram}
\end{figure}

As a result, urban and rural representations containing disproportionate spurious information levels will cause group-level model performance disparities in semantic segmentation. Note that there are other factors not uniformly distributed across urban and rural areas, such as the number of pixels by land object class. Since we focus on representation learning, we denote such factors as unmeasured confounders $U$. The problem is illustrated, in terms of causal relationships, in Figure~\ref{fig:diagram}. Different from methods which directly alter spurious features which are defined a priori, \textit{the goal here is to reduce the part of model representations that are correlated to the urban and rural split, an important delineation which has strong disparities globally}. That is, to obtain $\hat{X}_{robust}$ which promotes $\hat{Y}\perp S | \hat{X}_{robust}$, where the model prediction $\hat{Y}$ is independent to group discrepancy.

Accordingly, there is a need for model to (1) focus on robust and generalizable landscape features, and (2) capture local features of the image in the pre-training stage. With contrastive self-supervised pre-training as the framework, we propose an intervention algorithm to achieve the goals and promote urban-rural downstream segmentation equity.


\section{Methodology}
\label{sec:methodology}

\subsection{Datasets}
While several standard image datasets used in fairness studies exist, datasets with linked group-level properties, specifically, urbanization, for real-world satellite imagery are very limited. We identified two datasets which had or could be linked with urban/rural annotations for disparity analyses, collected from Asia and Europe respectively, and with different spatial resolutions: 

\textit {LoveDA ~\cite{wang2021loveda}} is composed of 0.3m spatial resolution RGB satellite images collected from three cities in China. Images are annotated at pixel-level into 7 land-cover object classes, also with a label based on whether they are from an urban or rural district. Notably, images from the two groups have different class distributions. For example, urban areas contain more buildings and roads, while rural areas contain larger amounts of agriculture ~\cite{wang2021loveda}. Moreover, it has been shown that model segmentation performances differ across urban and rural satellite images ~\cite{zhang2022segmenting}. We split the original images into 512$\times$512 pixel tiles, take 18\% of the data for testing, and for the rest, 90\% are for contrastive pre-training (5845 urban tiles and 5572 rural tiles)  and 10\% for fine-tuning the pre-trained representation to generate predictions.

\textit {EOLearn Slovenia ~\cite{Sinergise}} is composed of 10m spatial resolution Sentinel-2 images collected from whole region of Slovenia for the year 2017, with pixel-wise land cover annotations for 10 classes. We only use the RGB bands for the consistency with other datasets, remove images that have more than 10\% of clouds, and split images into 256$\times$256 pixel tiles to enlarge the training set. Labels are assigned by assessing if the center of each tile is located in urban boundaries or not (using urban municipality information\footnote{https://www.gov.si/en/topics/towns-and-protected-areas-in-slovenia/} and administrative boundaries from OpenStreetMap\footnote{https://www.openstreetmap.org/\#map=12/40.7154/-74.1289}). This process generates 1760 urban tiles and 1996 rural tiles in total. Similar to the LoveDA process, 18\% of the data are used for testing, and 90\% of the rest of the data are used for pre-training and 10\% for fine-tuning.

\subsection{Metrics}
The quality of representations learnt from self-supervised pre-training is usually evaluated by its transfer-ability to downstream tasks ~\cite{jing2020self, wang2021dense}. On the downstream semantic segmentation, we use Intersection-over-Union (IoU) as the accuracy metric, calculated using pixel-wise true positives ($TP$), false positives ($FP$), and false negatives ($FN$),
\begin{align*}
\text{IoU} := \frac{TP}{TP + FP + FN}. 
\end{align*}
Group accuracy for group $g^i$ is computed via the mean of class-wise IoUs (referred to as $\mu_{g^i}$). Model overall accuracy is the averaged group results (mIoU).

We use two fairness metrics: First, the group difference with regard to accuracy~\cite{raff2018gradient, gong2021mitigating, szabo2021tilted, zietlow2022leveling} (Diff). Diff for a 2-element sensitive attribute group $\{g^1, g^2\}$ is defined as:
\begin{align*}
\text{Diff $\{g^1, g^2\}$} := \frac{|\mu_{g^1}-\mu_{g^2}|}{\min\{\mu_{g^1}. \mu_{g^2}\}}. 
\end{align*}

Second, the worst group results (Wst), which is the lower group accuracy between urban and rural. This is motivated by the problem of worsening overall performance for zero disparity ~\cite{zhang2022improving}.

\subsection{Multi-Level Representation De-biasing}
\label{sec:multi-level}
The idea of constraining mutual information between representation and sensitive attribute, also referred to as bias, to achieve attribute-invariant predictions has multiple applications ~\cite{zhu2021learning, ragonesi2021learning, kim2019learning}, which all operate on a global representation $\mathbf{z} = F(\mathbf{d})$, output from image encoder $F$. However, invariance constraints only on the global output layer do not guarantee that sensitive information is omitted from representation hierarchies of intermediate layers or blocks in a model (herein we use the term ``multi-level representation'' for simplicity). As has been shown, the distribution of bias in terms of its category, number and strength is not constant across layers in contrastive self-supervised models ~\cite{sirotkin2022study}. Besides, layer-wise regularization is necessary to constrain the underlying representation space~\cite{jin2016collaborative,jiang2017learning,li2019learning}. Pixel-level image features in representation hierarchies are important ~\cite{o2020unsupervised, wang2021dense}, especially when transferring to dense downstream tasks such as semantic segmentation, where representations are aggregated at different resolution scales in order to identify objects in pixel space. Given the evidences in sum, we design a feature map based local mutual information estimation module and incorporate layer-wise regularization into the contrastive optimization objective.

To measure mutual information $MI(X, S)$ between local feature $X$ and the urban/rural attribute $S=\{s_0, s_1\}$, we adapt the concat-and-convolve architecture in ~\cite{hjelm2018learning}. Notating the $i^{th}$ layer as $li$, we first build a one-hot encoding map $\mathbf{c}^{li}$ for attributes $S$ whose size is same as the feature map $\mathbf{x}^{li}$ output by $li$, and channel is the size of $S$. For each $\mathbf{x}^{li}$, a $\mathbf{c}^{li}$ is built from the joint distribution of representation space $X$ and attribute space $S$, and the marginal distribution of $S$ separately, then the $\mathbf{c}^{li}$ built in the two ways are concatenated with $\mathbf{x}^{li}$ to form an ``aligned'' feature map pair, denoted as $P_{XS}(\mathbf{x}^{li} \mathbin\Vert \mathbf{c}^{li})$, and a ``shuffled'' feature map pair, denoted as $P_{X}P_{S}(\mathbf{x}^{li} \mathbin\Vert \mathbf{c}^{li})$. The mutual information between the aligned and shuffled feature map pairs will be estimated by a three-layer $1\times1$ convolutional discriminator $D_i$, using the JSD-derived formation ~\cite{hjelm2018learning}: 
\begin{align*}
MI_{JSD}(X^{li}; S) := E_{P_{XS}}[-\text{sp}(-D_i(\mathbf{x}^{li} \mathbin\Vert \mathbf{c}^{li}))] \\
-E_{P_{X}P_{S}}[\text{sp}(D_i(\mathbf{x}^{li} \mathbin\Vert \mathbf{c}^{li}))], 
\end{align*}
where sp$(a) = log(1+e^a)$, and $D_i$ uses separate optimization to converge to the lower bound of $MI_{JSD}$. 

\begin{figure}[t!]
  \centering
    \includegraphics[width=8.4cm]{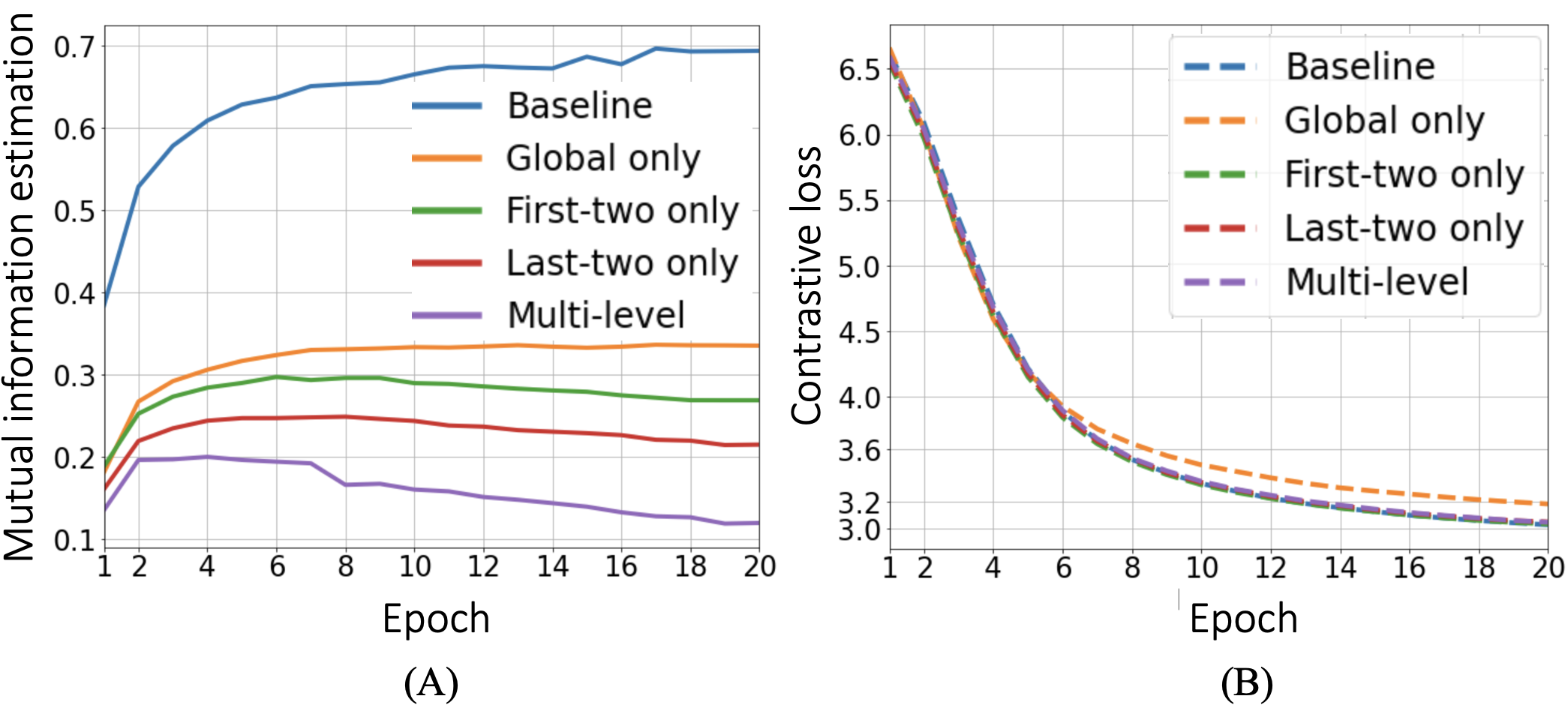}
    \caption{Bias accumulation during contrastive pre-training. (A) Sum of mutual information estimation, and (B) the contrastive loss of ResNet50 model with MoCo-V2 pre-training. The baseline method with no intervention (Baseline), regularizing only on the global feature vector (Global only), first two layers of feature maps (First-two only), last two layers of feature maps (Last-two only) all show bias residuals compared to the multi-level method proposed as part of FairDCL.}
    \label{fig:JSD}
\end{figure}

We empirically validate the necessity to apply multi-level constraints to reduce bias accumulation across layers. We run self-supervised contrastive learning on LoveDA data using MoCo-v2~\cite{chen2020improved} with ResNet50~\cite{he2016deep} as the base model. Simultaneous to model contrastive training, four independent discriminators are optimized to measure the mutual information $MI_{JSD}(X^{l1}; S), ...,  MI_{JSD}(X^{l4}; S)$ between representation output from the four residual layers and one output layer of ResNet50 and sensitive attributes: urban/rural. $MI_{JSD}$ are summed to measure the total amount of model bias for the data batch. 
As plotted in Figure~\ref{fig:JSD} (A), the baseline training without $MI_{JSD}$ intervention shows continually increasing and significantly higher bias than other methods as the number of epochs increase. Adding a penalty loss which encourages minimizing $MI_{JSD}$ only on the global representation or on subsets of layers both control bias accumulation, but their measurements are still high compared to multi-level, showing that global level regularization might remove partial bias but leave significant residual from earlier layers. 
The running loss during training indicates all methods' convergence (Figure~\ref{fig:JSD} (B)); mutual information constraints in latent space do not affect the contrastive learning objective.

\begin{figure}[h!]
  \centering
    \includegraphics[width=8.3cm]{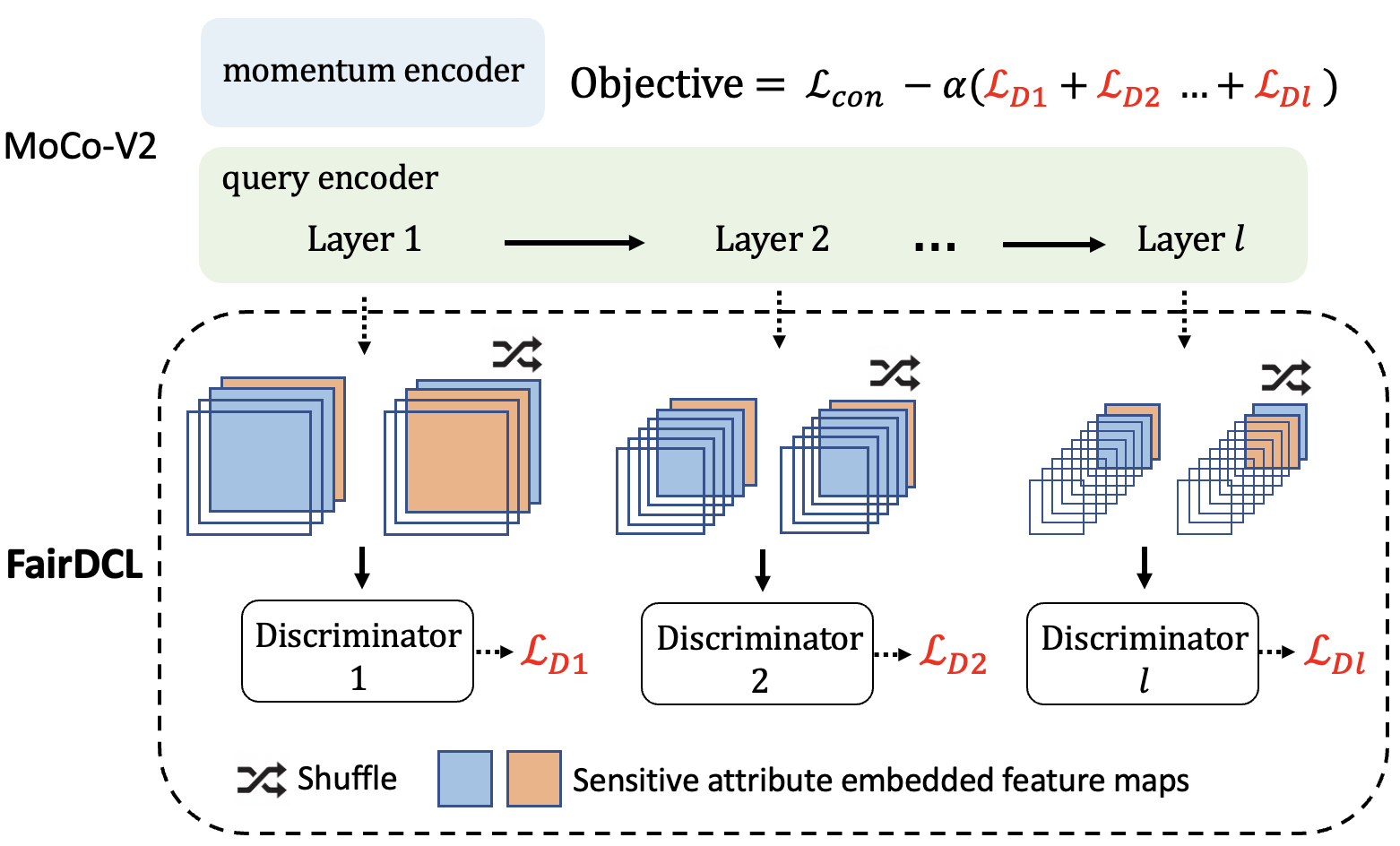}
    \caption{Overview of FairDCL. It captures spurious information $X_{spurious}$ learnt by urban/rural discriminators, and applies regularization on image representations at multiple levels. We build one-hot feature maps to encode urban/rural attribute and estimate mutual information by neural discriminators. Penalty loss $\mathcal{L}_{D_i}$ are computed accordingly and added into the final contrastive pre-training objective.  }
    \label{fig:arch}
\end{figure}

\setlength{\textfloatsep}{12pt}
\begin{algorithm}
\small
\newcommand{\vect}{\ensuremath{\mathbf{##1}}}

\textbf{for} each iteration a from $1$ to $E$ \textbf{do}: 

\quad \textbf{Image encoder forward propagation:}   

\quad $\mathbf{x}^{lN}, \mathbf{x}^{lN-1}, ..., \mathbf{x}^{l1} \leftarrow F(x)$ \quad $\triangleright$ $\mathbf{x}^{li}$ is the query representation output of the layer $li$  

\quad \textbf{Discriminators updating:}   

\quad \textbf{for} each round b from $1$ to $B$ \textbf{do}: 

\quad \quad \textbf{for} each discriminator $D_i$ \textbf{do}: 

\quad \quad \quad ${\mathcal{L}}_{Di} \leftarrow$ $D_i(\mathbf{x}^{li} \mathbin\Vert P_{XS} (\mathbf{c}^{li}), \mathbf{x}^{li} \mathbin\Vert P_S(\mathbf{c}^{li})) $ 

\quad \quad \quad $\triangleright$ Forward aligned and shuffled feature pairs 

\quad \quad \quad $W_{D_i} \leftarrow W_{D_i} - \eta \bigtriangledown  \mathcal{L}_{D_i}$ \hspace{0.3cm} $\triangleright$ Optimize $D_i$ 

\quad \textbf{Image encoder updating:}   

\quad $\mathbf{x}^{lN}, \mathbf{x}^{lN-1}, ..., \mathbf{x}^{l1}, q, k \leftarrow F(x)$ \quad $\triangleright$ $q, k$ is the query and key global representation  

\quad $\mathcal{L}_{con} \leftarrow q, k$ \quad \hspace{0.9cm} $\triangleright$ Compute contrastive loss 

\quad $\mathcal{L}_{D} \leftarrow \sum_{i=1}^{N}\mathcal{L}_{D_i} $ \hspace{1.2cm} \quad $\triangleright$  Compute MI loss 

\quad $W_F \leftarrow W_F - \eta ( \mathcal{L}_{con} - \alpha \mathcal{L}_D )$  $\triangleright$ Update encoders

 \caption{FairDCL. $F=\{l1,l2,..li..,lN\}$ is the contrastive learning encoder. $E$ is iterations per epoch; $B$ is discriminators updating rounds; $\eta$ is learning rate. $\alpha$ is regularization strength.}
\label{algo:b}
\end{algorithm}

\subsection{FairDCL Pipeline}
\label{sec:pipeline}
Figure~\ref{fig:arch} provides an overview of the proposed fair dense representations with contrastive learning (FairDCL) method and the training process (Detailed algorithm steps are in Algorithm~\ref{algo:b}). For each iteration of contrastive pre-training, latent space representation $\textbf{x}^{li}$ is yielded at layer $li$ of the encoder $F$. Layer discriminators $D_i$ are optimized by simultaneously estimating and maximizing $MI_{JSD}$ with the loss: 
\begin{equation}
\mathcal{L}_{D_i}(\textbf{x}^{li}, S; D_i) = -MI_{JSD}(\textbf{x}^{li}; S).
\end{equation}
Following ~\cite{ragonesi2021learning}, each $MI$ discriminator is optimized for multiple inner rounds before encoder weights get updated. More rounds are desirable for discriminators to estimate mutual information with increased accuracy, and based on resource availability, we set a uniform round number $B=20$. After discriminator optimization completes, one iteration of image encoder training is conducted wherein discriminators infer the multi-stage mutual information by loss in (1), and the losses are combined with the contrastive learning loss with a hyper-parameter $\alpha$ adjusting the fairness constraint strength. The final training objective is:
\begin{equation}
\mathcal{L}_F(X, S; D, F) = \mathcal{L}_{con} - \alpha(\sum_{li} \mathcal{L}_{D_i}(X^{li}, S; D_i)),
\end{equation}
With the training objective, the image encoder is encouraged to generate representation $X$ with high $\mathcal{L}_D$, thus low $MI_{JSD}$ (low spurious information). We apply FairDCL on the state-of-the-art contrastive learning framework MoCo-v2 ~\cite{chen2020improved}. The  loss used for learning visual representation is InfoNCE ~\cite{oord2018representation}: 
\begin{equation}
\mathcal{L}_{con}( F ) = -log \frac{\text{exp}(qk/\tau)}{\text{exp}(qk/\tau ) + \sum_j ( q\hat{k_j}/\tau )}.
\end{equation}
Here $F$ consists of a query encoder and a key encoder, which outputs representations $q$ and $k$ from two augmented views of the same image. $\hat{k_j}$ is a queue of representations encoded from different images in the dataset. $\mathcal{L}_{con}$ encourages the image encoder to distinguish positive and negative keys so it can extract useful visual representations.

\begin{table*}[h!]
  \small
  \begin{center}\fontsize{9}{11}\selectfont
    \setlength{\tabcolsep}{4.1pt}
    \scalebox{1}{
    \begin{tabular}{*{28}{c} | c}
        \toprule
                  &  & \multicolumn{3}{c}{LoveDA} & \multicolumn{3}{c}{Slovenia}  \\
                  \cmidrule(lr){3-5}
                  \cmidrule(lr){6-8}
                  &  Method & Diff($\downarrow$)& Wst($\uparrow$)& mIoU($\uparrow$) & Diff($\downarrow$)& Wst($\uparrow$)& mIoU($\uparrow$) 
                  \\
        \midrule
        \midrule
                  & Vanilla  & 0.165 ($\pm$  0.010) & 0.498 ($\pm$  0.004) & 0.539 ($\pm$  0.002) & 0.150 ($\pm$ 0.028) & 0.205 ($\pm$ 0.003) & 0.220 ($\pm$ 7e-4)  \\
           & GR   & 0.155 ($\pm$ 0.013) & 0.501 ($\pm$ 0.005) & \underline{0.540} ($\pm$ 0.002) & 0.128 ($\pm$ 0.023) & 0.208 ($\pm$ 6e-4) & \underline{0.222} ($\pm$ 0.002)   \\
           Moco-v2   & DI  & 0.144 ($\pm$ 0.009) & 0.499 ($\pm$ 0.004) & 0.535 ($\pm$ 0.004) & 0.136 ($\pm$ 0.010) & 0.208 ($\pm$ 0.005) & \underline{0.222} ($\pm$ 0.005)  \\  
                  & UnbiasedR  & 0.150  ($\pm$ 0.008) & 0.502 ($\pm$ 0.003) & \underline{0.540}  ($\pm$ 0.003) & 0.130 ($\pm$ 0.017) &  0.205 ($\pm$ 0.004) & 0.219 ($\pm$ 0.004)   \\ 
                  & FairDCL  & \textbf{0.127} ($\pm$ 0.005) & \textbf{0.508} ($\pm$ 0.002) & \underline{0.540} ($\pm$ 0.002) & \textbf{0.076} ($\pm$ 0.011) & \textbf{0.217} ($\pm$ 0.002) & \underline{0.225} ($\pm$ 0.003)  \\  
        \midrule
                  & Vanilla   & 0.154 ($\pm$ 0.013) & 0.503 ($\pm$ 0.004) & 0.542 ($\pm$ 0.002) & 0.122 ($\pm$ 0.017) & 0.207 ($\pm$ 0.006) & 0.219 ($\pm$ 0.005) \\
          & GR    & 0.148 ($\pm$ 0.012) & 0.498 ($\pm$ 0.006)  & 0.534 ($\pm$ 0.004) & 0.120 ($\pm$ 0.020) & 0.201 ($\pm$ 0.003)  & 0.216 ($\pm$0.002)   \\
          DenseCL  & DI   & 0.140 ($\pm$ 0.007) & 0.501 ($\pm$ 0.003) & 0.536 ($\pm$ 0.002) & 0.120 ($\pm$ 0.008) & 0.206 ($\pm$ 0.001) & 0.218 ($\pm$ 5e-4)   \\
                  & UnbiasedR  & 0.157 ($\pm$ 0.008)  & 0.495 ($\pm$ 0.005) & 0.534 ($\pm$ 0.003) & 0.128 ($\pm$ 0.014) & 0.206 ($\pm$ 0.004) & \underline{0.219} ($\pm$ 0.003)   \\
                  & FairDCL  & \textbf{0.108} ($\pm$ 0.008) & \textbf{0.518} ($\pm$ 0.003) & \underline{0.546} ($\pm$ 0.004) & \textbf{0.079} ($\pm$ 0.016) & \textbf{0.215} ($\pm$ 0.001) & \underline{0.223} ($\pm$ 0.003)  \\
        \bottomrule \\
    \end{tabular}
    }
    \caption{Downstream semantic segmentation results on LoveDA and Slovenia datasets, using an FCN-8s model with the backbone learnt with comparison pre-training methods. Our FairDCL shows consistent improvements on fairness metrics (Diff and Wst) (bold) over other de-biasing methods , also we do not see a decreased accuracy (mIoU) (underlined) than the vanilla baseline, on all datasets. Results and standard deviations are reported over 5 independent runs.}
    \label{tab:table1}
  \end{center}
\end{table*}

\textit{Generalizability to contrastive frameworks.} We note that the proposed locality-sensitive de-biasing scheme applying intervention on embedding space can be integrated with any state-of-the-art convolution feature extractors, thus has the potential to be further promoted with different contastive learning frameworks. Empirically, we experiment with DenseCL~\cite{wang2021dense}, which designs pixel-level positive and negative keys to better learn local feature correspondences. Since the method fills the gap between pre-training and downstream dense prediction, it is suitable as an alternative contrastive learning framework for our proposed method. The results are attached in the supplementary material.

\section{Experiments}
\subsection{Implementation Details}
\label{sec:implementation}
\textit{The first stage of contrastive pre-training.} The base model for the image encoders is ResNet50~\cite{he2016deep}. The mutual information discriminators $D_i$ are built with $1\times 1$ convolution layers (architecture details in supplementary material D). The contrastive pre-training runs for 10k iterations for each dataset with a batch size of 64. Data augmentations used to generate positive and negative image view pairs are random greyscale conversion and random color jittering (no cropping, flips or rotations in order to retain local feature information). Hyper-parameter $\alpha$, which scales the amount of mutual information loss $\mathcal{L}_{D}$ in the total loss, is set to 0.5. Adam optimizer is used with a learning rate of $10^{-3}$ and weight decay of $10^{-4}$ for both encoders and discriminators. 

\textit{Comparison methods} include state-of-the-art fair representation learning approaches: (1) gradient reversal training (GR)~\cite{raff2018gradient}, which follows the broad approach of removing bias or sensitive information from learnt representations by inverse gradients of attribute classifiers. This approach has been adapted to multiple image recognition tasks~\cite{zhang2018mitigating, wang2019balanced, wang2022fairness}. (2) Domain independent training (DI) which samples data with a consistent group attribute in each training iteration to avoid leveraging spurious group boundaries ~\cite{tsai2021conditional, wang2020towards}. (3) Unbiased representation learning (UnbiasedR)~\cite{ragonesi2021learning} which uses single-level de-biasing only for the global image representation. All
comparison methods use the same learning architectures,
and are trained with the same settings.

\textit{The second stage of semantic segmentation fine-tuning.} Following the protocal of previous work~\cite{wang2021dense, zhang2021unleashing, ziegler2022self}, we train a FCN-8s~\cite{long2015fully} model on top of the fixed ResNet50 backbone learnt from the pre-training stage for 60 epochs with a batch size of 16, and evaluate on the testing split for each dataset. We use cross-entropy (CE) loss as the training objective, and stochastic gradient descent (SGD) as the optimizer with a learning rate of $10^{-3}$ and a momentum of 0.9. The learning rate is decayed using a polynomial learning rate
scheduler implemented in PyTorch. Image data augmentations used in the fine-tuning include random horizontal/vertical flips and random rotations. For both stages, the dimension of the input image to the model is 512$\times$512$\times$3 where 3 indicates the RGB bands. NVIDIA RTX8000 GPU is used for training.

\subsection{Results}
\label{sec:results}
\subsubsection{Downstream Performances}
\label{sec:downstream_performance}
Table~\ref{tab:table1} summarizes model fine-tuning results pre-trained with baseline: vanilla MoCo-v2, and de-biasing methods: GR, DI, UnbiasedR, and FairDCL, on the two satellite image datasets.

\begin{figure}[t!]
  \centering
    \includegraphics[width=8.4cm]{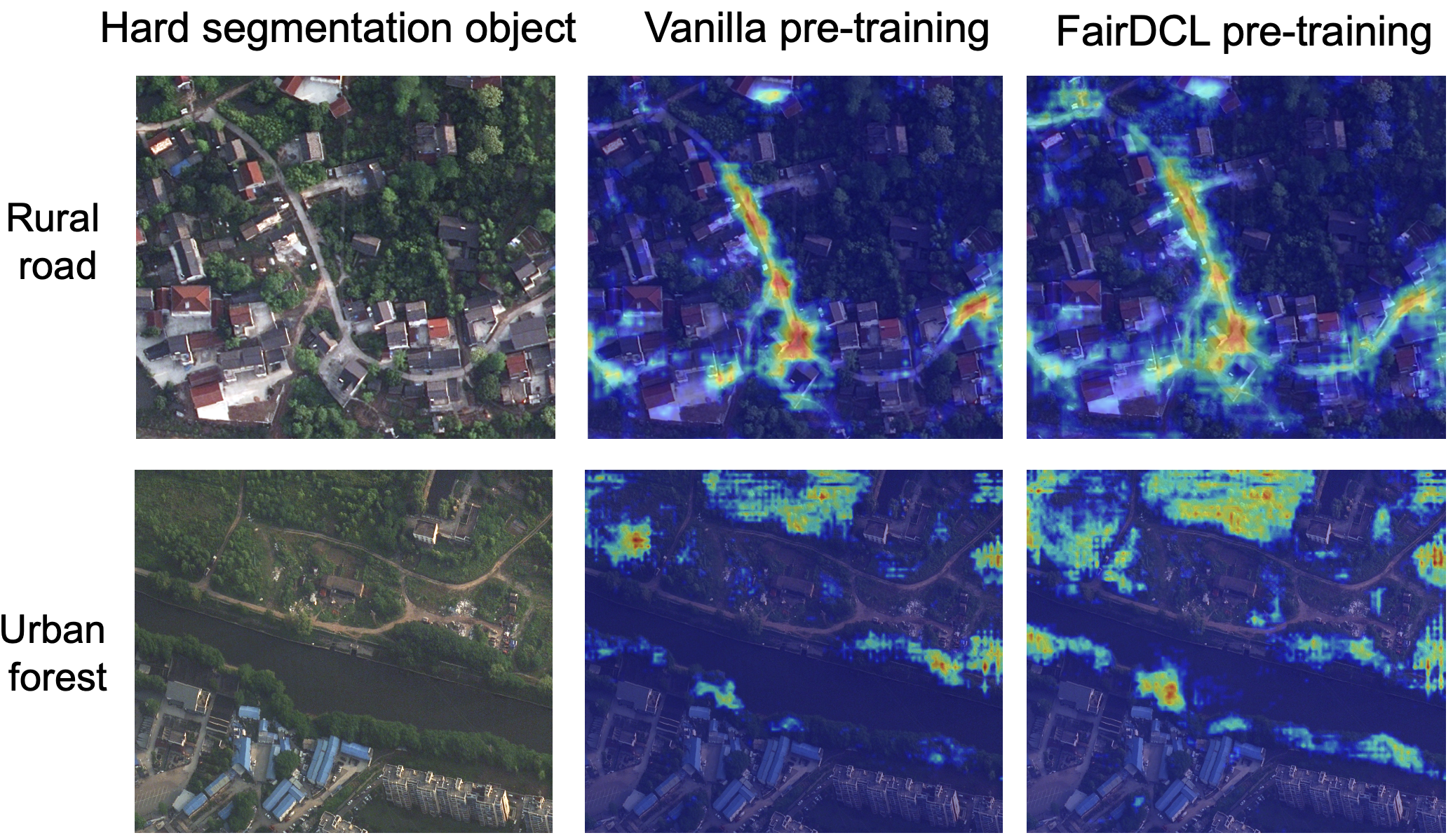}
    \caption{Class activation mapping. Detailed image locations that most impact model's prediction for ``road'' (first row) and ``forest'' (second row). FairDCL better recognizes land-cover segments particular to sensitive attributes.}
    \label{fig:activation}
\end{figure}

We first note that FairDCL consistently outperforms other approaches in terms of fairness, that it obtains the smallest cross-group difference (Diff) and highest worst group result (Wst). To reveal model decision process, we draw class activation maps for ``road'' and ``forest'' in Figure~\ref{fig:activation} using Grad-CAM~\cite{selvaraju2017grad}. Compared to the vanilla baseline, the model pre-trained with FairDCL better activates and recognizes tricky land-cover segments: curved part of roads (more seen in rural area), and sparse river-side forests (more seen in urban area). The method can learn robust representations that are generalizable to object shape and context variations, using the multi-scale representation regularization. Therefore, it reduces segmentation disparity caused by landscape discrepancy between groups. 

Another advantage of learning better features to ensure generalizability is no subsequent target task degradation, which is crucial for real-world applications. Importantly, FairDCL does not show degraded accuracy (the improved ``Diff" metric or ``Wst" metric does \textit{not} cause a worse ``mIoU" metric) on all cases. Obtaining comparable or better overall accuracy to Baseline demonstrates robustness in addition to disparity reduction. In contrast, DI allows image contrastive pairs only from a fraction of data which can discount model learning~\cite{wang2020towards}. The adversarial approach used in GR can be counter-productive if the adversary is not trained enough to achieve the infimum~\cite{moyer2018invariant}, which could all potentially degrade model quality for group equalization. Our adapted mutual information constraints use information-theoretic objectives, proved to be able to optimize without competing with the encoder so can match or exceed state-of-the-art adversarial de-biasing methods ~\cite{moyer2018invariant, ragonesi2021learning}. FairDCL further shows that applying the mutual information constraints on multi-level latent representations can better extend fairness to pixel-level applications, which outperforms the image-level only constraints used in UnbiasedR.

\subsubsection{Embedding Spaces}
\label{sec:embedding_space}

\begin{figure}[t!]
  \centering
    \includegraphics[width=8.4cm]{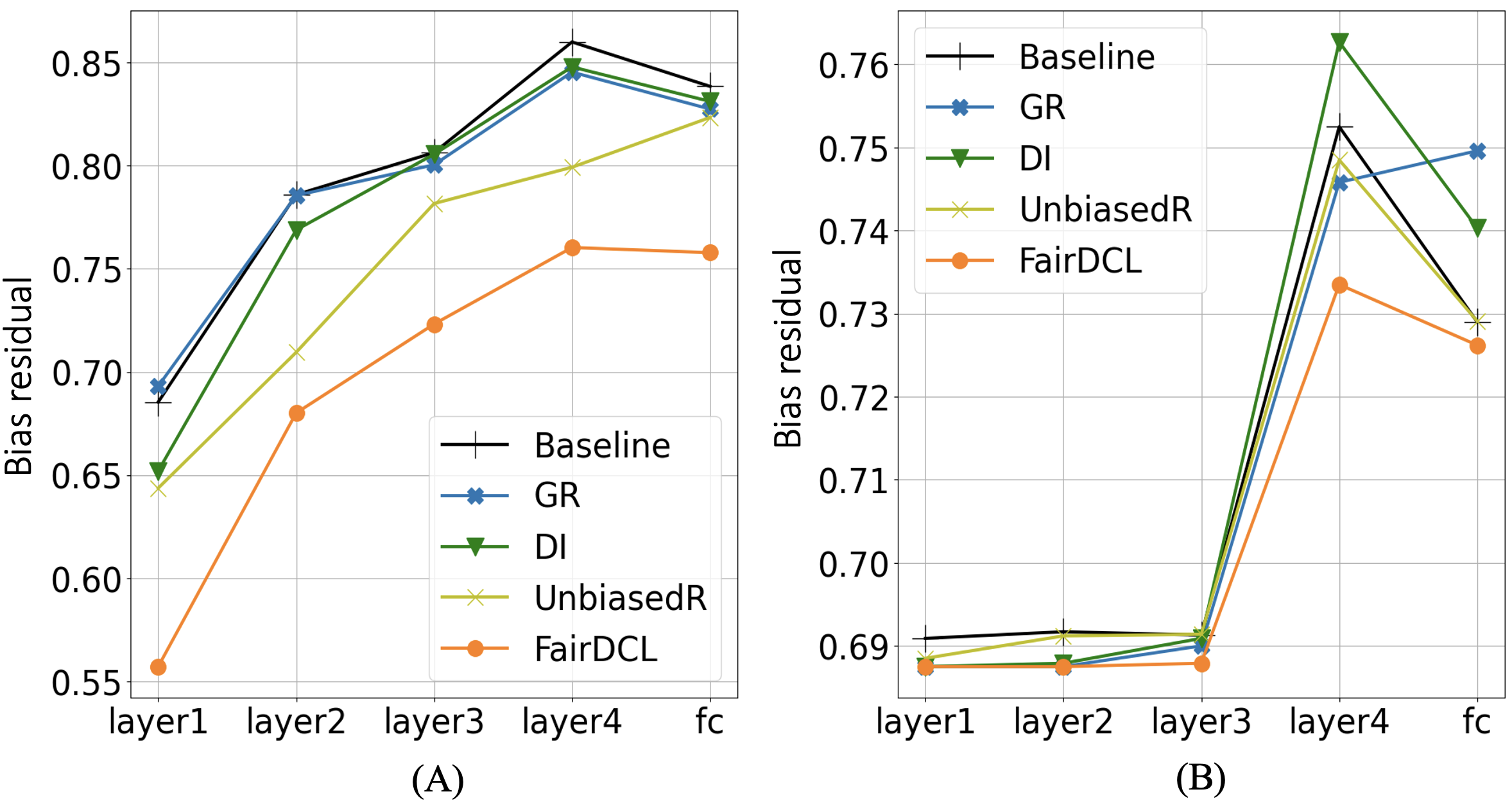}
    \caption{Linear separation evaluation: We train a linear neural layer on top of each level of representations output from different model layers. They include four residual modules (``layer1'' - ``layer4'') that encode intermediate representations and a global output layer (``fc'') that encodes the global representation. The linear layer is to classify sensitive attributes: urban/rural on (A) LoveDA dataset, and women/men on (B) MS-COCO dataset. \textit{Lower accuracy is good}: it indicates harder to predict sensitive attributes using the pre-trained representations, thus lower bias residual. }
    \label{fig:linear}
\end{figure}

To further trace how the image representations learnt with proposed method improves fairness, we analyze a linear separation property~\cite{reed2022self} on model embedding spaces. Specifically, we assess how well a linear model can differentiate urban/rural sensitive attributes using learnt representations. High separation degree indicates that the encoder model's embedding space and attribute are differentiable ~\cite{oord2018representation, chen2020simple, reed2022self}, which could be used as a short-cut in prediction and cause bias, thus is not desirable here. We freeze the trained ResNet50 encoder and use a fully connected layer on top of representation output from different layers for urban/rural attribute classification. Figure~\ref{fig:linear} (A) presents the classification score on urban/rural attribute on LoveDA: FairDCL obtains the lowest attribute differentiation results for all embedding stages and global stage of representation, indicating that the encoder trained with FairDCL has favorably learnt the least sensitive information at pixel-level features during the contrastive pre-training. 
Though we focus on satellite images, we check the method's generalizability to a different image domain by conducting contrastive pre-training on MS-COCO~\cite{lin2014microsoft}, a dataset commonly used in fairness studies~\cite{wang2019balanced, tang2021mitigating, wang2021gender}. Sensitive attribute gender, categoriezed as ``women" or ``men", is obtained from~\cite{zhao2021understanding}. There are 2901 images with ``women" and 6567 images with ``men" labels. Linear analysis results show that FairDCL again produces the desired lowest classification accuracies (Figure~\ref{fig:linear} (B)), but unlike in Figure~\ref{fig:linear} (A), it does not surpass the other comparison methods much. This is likely because while geographic attributes are represented at a pixel-level, human face/object, as a foreground, may not be represented through local features throughout the image, thus gender attributes are less pronounced as pixel-level biases in dense representation learning, which is what our proposed approach focuses on.

\begin{table}[h!]
  \small
  \begin{center}\fontsize{10}{11}\selectfont
    \setlength{\tabcolsep}{6pt}
    \scalebox{0.85}{
    \begin{tabular}{*{15}{c}}
        \toprule
                  \multicolumn{3}{c}{$\alpha$ = 0.1} & \multicolumn{3}{c}{$\alpha$ = 0.5} \\
                  \cmidrule(lr){1-3}
                  \cmidrule(lr){4-6}
                    Diff& Wst & mIoU & Diff& Wst& mIoU   \\
        \midrule
                   0.138 & 0.506 & \textbf{0.541} & 0.127 & \textbf{0.508} & 0.540 \\  
        \midrule 
                 \multicolumn{3}{c}{$\alpha$ = 1} & 
                  \multicolumn{3}{c}{$\alpha$ = 10} \\
                  \cmidrule(lr){1-3}
                  \cmidrule(lr){4-6}
                    Diff& Wst & mIoU & Diff& Wst& mIoU   \\
        \midrule
        0.127 & 0.506 & 0.538 & \textbf{0.126} & 0.506 & 0.538 \\
        \bottomrule
    \end{tabular}}
    \caption{Ablation study for discriminator weights. The fine-tuning result is shown for the representation pre-trained with $\alpha=0.1, 0.5, 1, 10$, of the proposed fairness objective. }
    \label{tab:table3}
  \end{center}
\end{table}

\begin{table}[h!]
  \small
  \begin{center}\fontsize{10}{11}\selectfont
    \setlength{\tabcolsep}{3pt}
    \scalebox{0.85}{
    \begin{tabular}{*{10}{c}}
        \toprule
                  & \multicolumn{3}{c}{Urban:68\% Rural:32\%} & &  \multicolumn{3}{c}{Urban:35\% Rural:65\%}  \\
                  \cmidrule(lr){2-4}
                  \cmidrule(lr){6-8}
                   Method& Diff($\downarrow$)& Wst ($\uparrow$)& mIoU($\uparrow$)  & & Diff($\downarrow$)& Wst ($\uparrow$)& mIoU($\uparrow$)  \\
        \midrule
        \midrule
                  Baseline  & 0.147 & 0.500 & 0.537 &   & 0.170 & 0.497 & 0.539  \\
            GR   & 0.154 & 0.497 & 0.535  & & 0.144 & 0.511 & \underline{0.547}   \\
                  DI & 0.145 & 0.498 & 0.534  & & 0.148 & 0.499 & 0.535  \\  
                  UnbiasedR & 0.146 & 0.500 & \underline{0.537}  & & 0.145 & 0.503 & \underline{0.539}    \\ 
                  FairDCL & \textbf{0.128} & \textbf{0.511} & \underline{0.543}  & & \textbf{0.122}  & \textbf{0.518} & \underline{0.549}  \\  
        \midrule \\
    \end{tabular}}
    \caption{Ablation study for unbalanced group data. The proportion of urban/rural samples in the pre-training data is adjusted such that one group has much less samples. FairDCL performs consistently with the data distirbution shifts.}
    \label{tab:table4}
  \end{center}
\end{table}

\subsubsection{Ablation Studies}

We perform an ablation study for hyper-parameter $\alpha$ which scales discriminator loss $\mathcal{L}_{D}$, thus the fairness regularization strength. The method is overall robust to the parameter (Table~\ref{tab:table3}); a large $\alpha$ like $10$ will not corrupt the downstream accuracy, and a small $\alpha$ like $0.1$ has lower fairness gain but still shows advantage over comparison methods in Table~\ref{tab:table1}. We select $\alpha=0.5$ for a balance. Furthermore, urban and rural groups have comparable training samples in earlier experiments (LoveDA is 5.8k and 5.5k, Slovenia is 1.7k and 1.9k for urban/rural). We intentionally reduce pre-training samples for certain groups to generate more unbalanced subsets. Shown in Table~\ref{tab:table4}, the proposed method shows robustness under the two less even group distributions.

\section{Discussion and Conclusion}
Among the broader fairness literature in visual recognition, work focusing on satellite imagery that depicts physical environments has been limited. This limitation is largely due to the difficulty in identifying population level biased landscape features. Also, disparity problems in satellite image recognition may get categorized as domain adaptation or transfer learning problems, other popular computer vision fields; though they share similar technical methods in bias mitigation and invariant representation learning, the specific objective of fair urban/rural satellite image recognition is to remove spatially disproportionate features that favor one subgroup over the others, beyond addressing covariate shift.

Here we define the scenario with a causal graph, showing that contrastive self-supervised pre-training can utilize spurious land-cover object features, thus accumulate urban/rural attribute-correlated bias. The biased image representations will result in disparate downstream segmentation accuracy between subgroups within a specific geographic area. Then, we address the problem via a mutual information training objective to learn robust local features with minimal spurious representations. Experimental results show fairer segmentation results pre-trained with the proposed method on real-world satellite datasets. In addition to disparity reduction, the method consistently avoids a trade-off between model fairness and accuracy. 

As future directions, a wider set of satellite datasets can be explored. The fairness analysis can be scaled to a greater number of attributes relevant to geography, in addition to urbanization. Methods to encode sensitive attributes in the model embedding space in addition to one-hot feature maps can also be explored. We encourage experimenting with different encoding mechanisms and mutual information estimators to improve fairness regularization performance across different real-world settings.

\section{Acknowledgments}
We acknowledge funding from NSF award number 1845487. We also thank Harvineet Singh and Vishwali Mhasawade for helpful
discussions.


\bibliography{aaai24}


\end{document}